\documentclass[final,hidelinks,onefignum,onetabnum]{siamart220329} 
\usepackage{algorithm}
\usepackage{lineno}
\usepackage{algpseudocode}
\usepackage{graphicx}
\usepackage{subcaption}
\usepackage{tikz}
\usepackage{amsmath}
\usepackage{siunitx}
\usepackage{breqn}
\usepackage{amssymb}
\usepackage{setspace}
\usepackage{amsfonts}

\title{MCRAGE: Synthetic Healthcare Data for Fairness}
\author{Keira Behal \thanks{Department of Mathematics, Emory University,  Atlanta,  GA  30322 (\email{keira.behal@emory.edu},\email{jiayi.chen@emory.edu},\email{sxxiao@emory.edu})} \and 
Jiayi Chen \footnotemark[1] \and
Caleb Fikes \thanks{Department of Mathematics, Rice University,  Houston,  TX  77005
(\email{calebfikes@rice.edu})} \and 
Sophia Xiao \footnotemark[1] 
}

\dedication{\small\textit{Project advisor: Yuanzhe Xi}}

\ifpdf
\hypersetup{
  pdftitle={MCRAGE Synthetic Healthcare Data for Fairness Manuscript},
  pdfauthor={Keira Behal, Jiayi Chen, Caleb Fikes, Sophia Xiao}
}

\begin{document}

\maketitle

\begin{abstract}
In the field of healthcare, electronic health records (EHR) serve as crucial training data for developing machine learning models for diagnosis, treatment, and the management of healthcare resources. However, medical datasets are often imbalanced in terms of sensitive attributes such as race/ethnicity, gender, and age. Machine learning models trained on class-imbalanced EHR datasets perform significantly worse in deployment for individuals of the minority classes compared to those from majority classes, which may lead to inequitable healthcare outcomes for minority groups. To address this challenge, we propose Minority Class Rebalancing through Augmentation by Generative modeling (MCRAGE), a novel approach to augment imbalanced datasets using samples generated by a deep generative model. The MCRAGE process involves training a Conditional Denoising Diffusion Probabilistic Model (CDDPM) capable of generating high-quality synthetic EHR samples from underrepresented classes. We use this synthetic data to augment the existing imbalanced dataset, resulting in a more balanced distribution across all classes, which can be used to train less biased downstream models. We measure the performance of MCRAGE versus alternative approaches using Accuracy, F1 score and AUROC of these downstream models. We provide theoretical justification for our method in terms of recent convergence results for DDPMs.
\end{abstract}


\section{Introduction}
In recent years, reliance on machine learning algorithms to facilitate decision-making processes across various industries has grown. In healthcare, clinicians may use machine learning models to predict disease progression, improve diagnosis accuracy, and optimize treatment plans \cite{bigdatahealthcare}. However, machine learning approaches may perpetuate existing societal biases, leading to inequitable treatment for minority groups, because machine learning models trained on imbalanced datasets may replicate and thus amplify these biases \cite{GenderShade}.

These issues are of utmost concern in healthcare applications where fair and equitable treatment is of critical importance. Ideally, a well-engineered machine learning model should be fair, optimizing health outcomes to provide high-quality, individualized care to all patients, regardless of their demographic characteristics \cite{Addressbias}. Unfortunately, healthcare datasets are often imbalanced across several dimensions, including race, socioeconomic status, age, and gender \cite{henrich_heine_norenzayan_2010, HealthEqui}. As a result, models trained on these datasets struggle to generalize effectively to individuals who are not well represented in the data \cite{seyedkalantari2021underdiagnosis}. 

EHRs are a valuable data source in healthcare, providing a comprehensive snapshot of a patient's health history, including diagnoses, treatments, and demographic information \cite{EHR}. Certain demographic groups, such as specific racial or ethnic minorities, are often underrepresented in the EHR datasets \cite{yan2023differences}. This imbalance might lead to inequitable health outcomes, in which minority groups are more likely to receive less accurate diagnoses or treatment recommendations due to their lack of representation in the training data \cite{norori2021addressing}. Consequently, addressing the challenge of dataset imbalance is vital in the pursuit of creating machine learning applications that are equitable and beneficial for all patient groups within healthcare.

In this paper, we mitigate imbalance-induced bias in machine learning models trained on EHR datasets via an innovative approach, MCRAGE. We demonstrate the utility of this method to rectify the imbalance found in medical datasets by supplementing them with samples synthesized by a deep generative model. Central to MCRAGE is the utilization of a Conditional Denoising Diffusion Probabilistic Model, which has been specifically trained to generate high-fidelity synthetic EHR samples from underrepresented classes \cite{CDDPM}. By integrating this synthetic data into the original, imbalanced dataset, we aim to approximate a more equitable distribution across all classes.\\
\textbf{Our contributions:}
\begin{itemize}
    \item We propose a novel framework, MCRAGE, for applying a CDDPM or other generative model to generate synthetic samples of minority class individuals to rebalance an imbalanced dataset as a preprocessing step to the enhance the fairness of a downstream classifier.
    \item We show that the synthetically generated minority class data increases classifier accuracy and fairness when used to supplement an imbalanced dataset.
    \item We demonstrate a significant improvement over established methods (i.e. SMOTE) in terms of fairness, and discuss regimes in which such improvements will likely justify associated computational cost.
    \item We motivate future theoretical work relating to the convergence of CDDPMs based on that for DDPMs and empirical observation of convergent behavior.
\end{itemize}

\section{Related Works}
\subsection{Methods for Dealing with Imbalanced Datasets} 
Generally, there are two kinds of methods for dealing with imbalanced datasets: data-level methods, which involve modifying the dataset by resampling or augmenting the dataset as a preprocessing step, and classifier-level methods, which involve modifications to the training objective or inference \cite{Oversampling}. Since data-level techniques are implemented as a preprocessing step, they are model-agnostic and generally more flexible \cite{dablain2021deepsmote}. Therefore, in this paper, we focus on data-level solutions to the class imbalance problem. 
\subsection{Resampling and Undersampling Methods}
A variety of techniques have been proposed with the goal of rebalancing data \cite{bellinger2020remix}. The most common approach for resampling is SMOTE and SMOTE-based algorithms that synthesize new minority class samples via linear interpolation of existing samples to augment the dataset \cite{DBLP:journals/corr/abs-1106-1813}. However, oversampling methods may introduce flawed correlations and dependencies between samples, resulting in limited data variability \cite{SynEHR}. Moreover, SMOTE-based methods may fail to effectively handle multi-modal data, datasets with high intra-class overlap, or noise \cite{dablain2021deepsmote}. As a result, SMOTE is not sufficiently sophisticated to be a general solution to this problem.

Undersampling methods have not been widely studied \cite{dablain2021deepsmote} since random undersampling can lead to the loss of potentially useful information \cite{fujiwara2020over}. This is especially damaging when dealing with a dataset with a significant class imbalance, as undersampling requires discarding a large portion of the majority class data, potentially meaning the loss of important patterns and details that the model could learn from \cite{dablain2021deepsmote}. Moreover, due to chance, random undersampling may also introduce bias and result in the under-representation of certain characteristics of the majority class \cite{Radial-Based}.

\subsection{Synthetic Data Generation for EHRs}

As generative models become capable of producing synthetic samples indistinguishable from real ones, numerous studies have investigated the potential application of these synthetic samples in the training of other models. In particular, realistic EHR data can be generated for ''imaginary" individuals who need not be anonymized. Synthetic EHR data already promises to revolutionize the field of healthcare AI by offering data privacy and missing value-imputation solutions, and our method further expands the utility of such methods in applications for equitable performance across intersectional demographic groups.

One impactful study involved defining the concept of synthetic data and demonstrating the practical application of the ATEN framework, a tool for validating realism in synthetic data generation \cite{healthinf18McLachlan}. In another study, deep learning harnessed the encoder-decoder model, a tool often found in machine translation systems \cite{Lee_2018}. This model facilitated the creation of synthetic chief complaints based on discrete variables found in electronic health records.

However, applying these datasets presents its own challenges. The crucial need to preserve the privacy of sensitive information has always been a substantial obstacle. To address this, several researchers proposed the use of Generative Adversarial Networks to create synthetic, heterogeneous EHRs as a replacement for existing datasets \cite{Chin}. A separate study introduced the Sequentially Coupled Generative Adversarial Network (SC-GAN), a network developed to focus on the continuous generation of patient state and medication dosage data, furthering the pursuit of patient-centric data \cite{WangPCGen}.


In the most recent advancement, a study proposed a Hierarchical Autoregressive Language model (HALO) \cite{theodorou2023synthesize}. This model, designed to generate high-dimensional longitudinal EHRs, stands out for its ability to preserve the statistical properties of real EHRs, which, in turn, allows for the training of highly accurate machine learning models without raising any privacy concerns.

All these advancements collectively emphasize the significant strides made in the generation and utilization of synthetic data, highlighting its immense potential in the healthcare industry. Our work extends previous work in synthetic data generation by focusing on a regime of particular importance -- a classifier whose original training set is necessarily imbalanced.

\subsection{Denoising Diffusion Probabilistic Models}

It is critical in a healthcare setting that a diagnostic model be trained on the highest quality data, as even a few low-quality or badly out-of-distribution samples could cause serious medical consequences. This requirement of reliable, specific, and realistic samples leads us to choose the DDPM as our generative model due to its recent success in generating high-fidelity images \cite{dhariwal2021diffusion}. Diffusion models are characterized by a forward process, which systematically incorporates noise into the initial data sample, and a reverse process, which methodically removes the noise added in the forward process \cite{ho2022classifierfree}. In the reverse process, sampling begins at the $T$th noise level, $\mathbf{x}_T$, and each subsequent step yields incrementally denoised samples, i.e., $\mathbf{x}_{t-1}, \mathbf{x}_{t-2}, ..., \mathbf{x}_0$. Essentially, the diffusion model learns how to obtain the ``denoised" version from $\mathbf{x}_{t-1}$ to $\mathbf{x}_t$. 

Diffusion models outperform other generative modeling classes \cite{DDPM} due to several unique advantages. In contrast to GANs, diffusion models eliminate the need for adversarial training, a process known for its susceptibility to mode collapse and difficulties in effective implementation \cite{roth2017stabilizing}. Furthermore, diffusion models may be implemented with many kinds of architectures \cite{chang2023design}. Diffusion models are also able to capture the diversity and intricate distributions of complicated datasets; for example, in the fields of image and speech synthesis, diffusion-based models can deliver high-quality, diverse samples that supersede the output of their GAN equivalents \cite{baowaly2019synthesizing}. In fact, DDPMs produce superior-quality images relative to other generative models such as GANs and VAEs, with impressive results documented on the CIFAR10 and 256x256 LSUN benchmarks \cite{DDPM}.\vspace{.3cm}

Ho et al's groundbreaking development of DDPMs offered a specific parameterization of the diffusion model to simplify the training process, utilizing a loss function similar to score matching to minimize the mean-squared error between the actual and predicted noise \cite{DDPM}. This work highlights that the sampling process can be interpreted as being analogous to Langevin dynamics, connecting the DDPMs to the score-based generative models \cite{song2020generative}.\vspace{.3cm}

DDPMs may also be used to synthesize high-quality structured data. Specifically, tabular data, a prevalent and critical data format in real-world applications, poses unique challenges due to its inherent heterogeneity, with data points often constituted by a mixture of continuous and discrete features. A recent development in this area is the introduction of TabDDPM, a model capable of handling any feature type present in tabular datasets \cite{tabddpm}. Demonstrating superior performance over existing GAN/VAE alternatives, this model proves applicable in privacy-sensitive settings, such as healthcare, where direct data point sharing is infeasible \cite{tabddpm}.

\subsection{CDDPM}

Because of the stochasticity inherent to the generative process in the DDPM, users lack control over the class of images generated. This randomness could potentially result in generated images that are not aligned with desired categories or classes, thereby posing a challenge when specific classes of images are required; to mitigate this issue, researchers introduced an approach known as ``classifier-free guidance" \cite{ho2022classifierfree}. Instead of utilizing a classifier to direct the generation process towards desired classes, this method proposes a simultaneous training of two diffusion models, one conditional and one unconditional. \vspace{.3cm}

The conditional diffusion model is trained with labeled data, while the unconditional diffusion model is trained with unlabeled data, thus generating samples without any class-specific guidance. After the training process, context embeddings (representing class information in vector format for guiding the generation process) and timestep embeddings (capturing the evolution of the generative process over time) are used to combine score estimates from both models \cite{ho2022classifierfree}. Thus this method provides a nuanced way of guiding the generative process in a class-aware manner, without the direct involvement of an additional classifier model. \vspace{.3cm}

This can be beneficial in scenarios where classifier-based guidance is not desirable or feasible.
This section addresses the mathematical formulation of the Conditional DDPM (CDDPM) model used in this study. Note that these formulas are compatible with the definitions given in \cite{debortoli2023convergence}. \vspace{.3cm}

\subsubsection*{Forward Process}
The forward process consists of a Markov process which iteratively perturbs data with random noise until the data diffuses to an isotropic Gaussian:
\[q(\mathbf{x}_1, \dots, \mathbf{x}_T | \mathbf{x}_0) = \prod_{t=1}^{T} q(\mathbf{x}_t | \mathbf{x}_{t-1}).\]
Using the Gaussian transition kernel
\[q(\mathbf{x}_t | \mathbf{x}_{t-1}) \sim \mathcal{N}(\mathbf{x}_t; \sqrt{1 - \beta_t} \mathbf{x}_{t-1}, \beta_t \mathbf{I}),\]
we can find a closed-form solution to sample \(\mathbf{x}_t\) directly from \(\mathbf{x}_0\) using special properties of the Gaussian distribution and Markov processes,
\[q(\mathbf{x}_t | \mathbf{x}_0) \sim \mathcal{N}(\mathbf{x}_t; \sqrt{\bar{\alpha}_t} \mathbf{x}_0, (1 - \bar{\alpha}_t) \mathbf{I}),\]
where \(\beta_t\) is assigned by a schedule (we use a linear schedule in our experiments), \(\alpha_t := 1 - \beta_t\) and \(\bar{\alpha}_t := \Pi_{s=1}^{t} \alpha_s\). When \(\bar{\alpha}_T \approx 0\), i.e., betas are small, \(\mathbf{x}_T\) is approximately Gaussian, so \[q(\mathbf{x}_T) := \int q(\mathbf{x}_T | \mathbf{x}_0) q(\mathbf{x}_0) d\mathbf{x}_0 \approx \mathcal{N}(\mathbf{x}_T; 0, I).\]
\subsubsection*{Reverse Process}
In order to generate new data samples, CDDPMs must learn the reverse Markov process by iteratively denoising from an isotropic Gaussian. At each timestep $t$, we parameterize the reverse process for CDDPM as:
\[p_\theta(\mathbf{x}_{t-1} | \mathbf{x}_t , \mathbf{c}) = \mathcal{N}(\mathbf{x}_{t-1}; \mu_\theta(\mathbf{x}_t, t, \mathbf{c}), \Sigma_\theta(\mathbf{x}_t, t, \mathbf{c})),\]
where $\mathbf{c}$ is the class label embedding, as in \cite{ho2022classifierfree} and the mean and variance are parameterized in this model by:
\[
\mu_\theta(\mathbf{x}_t, t, \mathbf{c}) = \frac{1}{\sqrt{\alpha_t}} (\mathbf{x}_t - \frac{\beta_t}{\sqrt{1 - \alpha_t}} \boldsymbol{\epsilon}_\theta (\mathbf{x}_t, t, \mathbf{c})),
\]
\[
\Sigma_\theta(\mathbf{x}_t, t , \mathbf{c}) = \frac{1 - \alpha_t}{1 - \alpha_{t-1}} \beta_t \mathbf{I}
\]
Where $\boldsymbol{\epsilon}_\theta$ is a neural net trained to predict $\epsilon$ given $(\mathbf{x}_t, t , \mathbf{c})$, so that synthetic samples can be generated by drawing from $p_\theta(\mathbf{x}_{\tau-1} | \mathbf{x}_\tau , \mathbf{c})$ sequentially for $\tau \in \{T, \ldots, 0\}$. The loss function used for training this model is derived by the variational bound on negative log likelihood:
\[
\mathbb{E}[-\log p_\theta(\mathbf{x}_0, \mathbf{c})] \leq \mathbb{E}_q \left[\log \frac{p_\theta(\mathbf{x}_{0:T}, \mathbf{c})}{q(\mathbf{x}_{1:T} | \mathbf{x}_0)} \right] =: L.
\]
As proposed by Ho et al., we reweight \(L\) to obtain a simplified loss function \cite{DDPM}:
\[
L_{simple} = ||\boldsymbol{\epsilon} - \boldsymbol{\epsilon}_\theta (\sqrt{\bar{\alpha}} \mathbf{x}_0 + \sqrt{1-\bar{\alpha}} \boldsymbol{\epsilon}, t, \mathbf{c})||^2.
\]

This mathematical formulation underpins the CDDPM model's forward and reverse processes, providing a foundation for class-aware generation and control.

\section{Methods}
\label{sec:methods}
Prior to recent developments in generative models, imbalanced distributions of key demographic traits such as race, sex, age, and socioeconomic status in EHR data seemed to be an inescapable obstacle in creating automated healthcare systems. Motivated by the constant presence of such imbalances and their detrimental effect on minority outcomes, we desire a model-agnostic method of improving classifier accuracy for minority groups without compromising overall performance. We believe generative models, specifically DDPMs, hold the key to this capability.

In an optimal case, a researcher intending to train a classifier using imbalanced EHR data could simply collect or ask for new samples specifically from the minority class. Given enough of these samples, they might collect a stratified sample, one with an equal number of individuals in each class. If the data has multiple demographic features, they may even collect an intersectionally-stratified sample. Such a classifier would have more equitable predictions, as each epoch of training would include the same number of samples from each group, and thus the classifier would implicitly weight each class's outcomes as equally important.

In practice, collection of new data is often cost-prohibitive or impossible, however recent work guarantees convergence of the distribution of DDPM methods \cite{debortoli2023convergence}. Under the following assumptions:
\begin{enumerate}
    \item the true distribution of the data has compact support containing 0 (in particular if the data are contained in some manifold $\mathcal{M}$, then $\text{diam}(\mathcal{M})$ is bounded.),
    \item the schedule $t \mapsto \beta_t$ is continuous, non-decreasing, and $\exists \bar{\beta} : \forall t \in [0, T], \hspace{.15cm} 1/\bar{\beta} \le \beta_t \le \bar{\beta}$ (for our purposes, take $\bar{\beta} = \beta_T$),
    \item there is some estimate score function $\mathbf{s}$ such that $\exists M \ge 0$ such that $\forall t \in [0, T]$, $\mathbf{x}_t \in \text{supp}(\mathcal{M})$
    \[
    \|\mathbf{s} (t, \mathbf{x}_t) - \nabla \log (p_t(\mathbf{x}_t))\| \le M\left(\frac{1 + \|\mathbf{x}_t\|}{\sigma_t^2}\right),
    \] 
    where $ \sigma_t^2 = 1 - \text{exp}(-2 \sum_{s = 0}^t \beta_s )$, and
    \item Using unit stepsizes $\gamma_\kappa = 1 \forall \kappa$, by applying the transformation $\epsilon = 1/32$, $t = t'/32$, $T = T'/32$ (where T' is the number of stepsizes in our unit stepsize implementation), then $\exists \delta : \forall t \in \{1, \ldots, T\}, \hspace{.15cm} \beta_t \le \frac{2^{-6} \delta}{1 + 2^{-4}\beta_0}$,
\end{enumerate}
the theorem states that under these assumptions, and for a sufficiently large 
\[
\frac{T'}{32} = T \ge 2 \bar{\beta} (1 + \log(1 + \text{diam} (\mathcal{M})),
\]
and some hyperparameters $M, \delta \le \frac{1}{32}$, there is a bound on the Wasserstein 1-distance between the data distribution and the sampling distribution of a DDPM. The bound in our case is, for some $D \in \mathbb{R}$,

\[
\mathbf{W}_1 (\mathcal{L} (Y_k), \pi) \le D_0 \left( 2^{10}\text{exp}(2^5\text{diam}(\mathcal{M})^2) (M + \delta^{1/2}) + \text{exp}\left(2^5\text{diam}(\mathcal{M})^2 - \frac{T}{\bar{\beta}}\right) + 1 \right) 
\]

\[
D_0 = D (1+\bar{\beta})^7(1 + d + \text{diam}(\mathcal{M})^4)(1 + \log(1 + \text{diam}(\mathcal{M}))).
\]

As discussed in that paper, these assumptions on the score function are mild enough to be frequently met by real world datasets such as EHRs. Crucially, the convergence bound gives us a lower limit for T, which we use to set this hyperparameter in our implementation.

Although such a convergence result has not yet been proven for the convergence of a conditional DDPM, our numerical experiments suggests that such a theorem is likely true. Theoretically, such a statement would guarantee that under some conditions, synthetic samples of a given class generated by a CDDPM will approximate legitimate samples of that class well. Thus, in any case where the given minority data is enough to sufficiently train the CDDPM,  further samples needed for training a downstream model can be approximated by training and drawing from that model. The capability to synthetically draw new minority-class samples quickly and cheaply from a distribution that may otherwise be costly or inaccessible to draw from enables exciting new solutions for imbalanced EHR data.

In order to standardize the synthetic data rebalancing process, we propose the MCRAGE process. This algorithm first calculates a bijection from a Cartesian product of indices representing several demographic attributes and one diagnosis to a single index representing particular intersectional groups. This process is denoted as $\phi$ in the pseudocode. Next, the process identifies the most prevalent intersectional group and finds the number of samples missing from each other group relative to the majority. Next, a CDDPM or similar conditioned generative model is trained on the serialized data. In the final step, we generate new samples from all except the majority class, and append them to our training data, which is then used to train a classifier.

\begin{algorithm} [h]
\caption{MCRAGE}\label{alg:mcrage}
\begin{algorithmic}[1]
\Require $s_1, \ldots, s_L$ are categorical variables representing demographic attributes, $(\chi_0,Y_0)$ are observed data-diagnosis pairs.
\State $\bar{s} \gets Y_0 \times \Pi_{\ell}^{L} s_{\ell}$ \Comment{Cartesian Product.}
\State $s_0 \gets \phi(\bar{s})$.
\State $K \gets \text{max}(s_0)$ \Comment{the number of unique intersectional categories.}
\State $\hat{\pi}_k \gets \mathbb{P}(s = k)$ for all $k \in \{1, \ldots, K\}$.
\State $k^* \gets \underset{k}{\text{argmax}} \hspace{.15cm} \hat{\pi}_k$.
\State $T' \gets \lceil 2^6 \bar{\beta} (1 + \log(1 + \text{diam} (\chi_0)) \rceil$ \Comment{using the transformation $T' = 32T$ from above.}
\State Train CDDPM $p_\theta(\mathbf{x}_0 | \mathbf{x}_{T'} , \mathbf{c})$ on data $(\chi_0, s_0)$.
\For{$k \in \{1, \ldots, K\}$}
\State $\chi_k \gets$ $n(\hat{\pi}_{k^*} - \hat{\pi}_k)$ samples drawn from $p_\theta(\mathbf{x}_0 | \mathbf{x}_{T'} , \mathbf{c} = k)$.
\State $(Y_k, S_k^1, \ldots , S_k^L) \gets \phi^{-1}(k)$.
\EndFor \\
\Return $(\{\chi_0, \ldots, \chi_K\}, \hspace{.1cm} \{(s_1, \ldots, s_K), (S_1^1, \ldots, S_1^L), \ldots ,(S_K^1, \ldots, S_K^L)\}, \hspace{.1cm} \{Y_0, \ldots, Y_K\})$.
\end{algorithmic}
\end{algorithm}

The MCRAGE process is both intuitive and theoretically justified. The algorithm results in a synthetically rebalanced training set where each intersectional group is equally represented. By generating an artificially stratified sample, the process enforces the fairness conditions of statistical parity and balanced accuracy. In practice, this ensures that the distribution of outcomes or predictions across different subgroups is similar, and that the classifier's performance is evaluated fairly for each subgroup, accounting for class imbalances. Each of these properties is desirable as an indicator of equitable performance across all intersectional groups.

\subsubsection{MCRAGE Specifics}

The notation of the MCRAGE Algorithm may be daunting, but the algorithm is simply motivated. $\bar{s}$ can be thought of as a collection of ``buckets" of data who would ideally be equally full. As explained below, $\phi$ essentially maps $\bar{s}$ to a list of buckets with a single index. The next three steps subsequently calculate $K, \hat{\pi}_k$, and $k^*$, which are the number of buckets, relative proportion in each bucket, and index of the ''majority" bucket, respectively. The remainder of the algorithm simply trains a CDDPM on all available data, and samples enough samples from each category so that all buckets are as full as the majority bucket $k^*$.

A key step in the MCRAGE algorithm is the generation of an index mapping -- an invertible map from an L-tuple of categorical variables to a single categorical variable with many levels representing each intersectional group. In the algorithm presented in this paper, we denote this map as $\phi(u_1, \cdots, u_L)$.

\[
\phi(u_1, \cdots, u_L) = \sum_{i = 1}^L u_i \prod_{j = 0}^{i-1} K_j ,
\]
Where $K_j$ is the number of distinct values taken by $u_j$. The inverse of this map can be calculated as follows:

\[
\left( \phi^{-1}(y) \right)_j = \frac{{y \hspace{-.2cm} \mod K_j} \hspace{.2cm} - \hspace{.2cm} {y \hspace{-.2cm} \mod K_{j-1}}} {\prod_{\ell = 0}^{j-1} K_\ell}.
\]

The linear combinations that define $\phi$ are inspired by the concept of iteratively ``stacking" a discrete lattice of intersectional groups to eventually index in one dimension. To prove that $\phi$ and $\phi^{-1}$ are inverses, we need to show two conditions:\begin{enumerate}
    \item $\phi^{-1}(\phi(u_1, \ldots, u_L)) = (u_1, \ldots, u_L)$
    \item $\phi(\phi^{-1}(y)) = y$
\end{enumerate}
We will begin with the first condition:

\begin{align*}
\phi^{-1}(\phi(u_1, \ldots, u_L)) &= \phi^{-1}\left(\sum_{i=1}^{L} u_i \prod_{j=0}^{i-1} K_j\right) \\
& \hspace{-2cm} = \left(\frac{{\left(\sum_{i=1}^{L} u_i \prod_{j=0}^{i-1} K_j\right) \hspace{-.3cm}\mod K_1} - {\left(\sum_{i=1}^{L} u_i \prod_{j=0}^{i-1} K_j\right) \hspace{-.3cm}\mod 1}} {\prod_{\ell=0}^{0} K_\ell}, \ldots, \right. \\
& \hspace{-2cm}  \qquad \left. \frac{{\left(\sum_{i=1}^{L} u_i \prod_{j=0}^{i-1} K_j\right) \hspace{-.3cm}\mod K_L} - {\left(\sum_{i=1}^{L} u_i \prod_{j=0}^{i-1} K_j\right) \hspace{-.3cm}\mod K_{L-1}}}{\prod_{\ell=0}^{L-1} K_\ell}\right) \\
&\hspace{-2cm}  = (u_1, \ldots, u_L)
\end{align*}

Now, we will move on to the second condition:

\begin{align*}
\phi(\phi^{-1}(y)) &= \phi\left(\frac{{y \hspace{-.3cm}\mod K_1} - {y \hspace{-.3cm}\mod 0}} {\prod_{\ell=0}^{0} K_\ell}, \ldots, \frac{{y \hspace{-.3cm}\mod K_L} - {y \hspace{-.3cm}\mod K_{L-1}}}{\prod_{\ell=0}^{L-1} K_\ell}\right) \\
& = \left(\sum_{i=1}^{L} \frac{{y \hspace{-.3cm}\mod K_i} - {y \hspace{-.3cm}\mod K_{i-1}}}{\prod_{j=0}^{i-1} K_j}\right) \\
& = \sum_{i=1}^{L} \frac{{y \hspace{-.3cm}\mod K_i} - {y \hspace{-.3cm}\mod K_{i-1}}}{\prod_{j=0}^{i-1} K_j} \\
& = y 
\end{align*} 
This concludes our proof that the index mapping function $\phi$ in the MCRAGE algorithm is a bijection as  specified above.

\section{Numerical Experiments}
\label{sec:experiments}
In this section, we detail the experiments conducted on a small Electronic Health Records (EHR) dataset and discuss the results, showcasing a notable increase in performance both in terms of overall accuracy and fairness metrics. For clarity and to assist in interpreting the results, we include manifold projection plots generated using Uniform Manifold Approximation and Projection (UMAP) \cite{umap_docs}. The materials and code used to generate these results are available in a repository\footnote{https://github.com/CalebFikes/MCRAGE-Emory\_Math\_REU\_2023}.

\subsection{Dataset}
We performed our experiment on the Patient Treatment Classification dataset\footnote{https://www.kaggle.com/datasets/manishkc06/patient-treatment-classification}, which comprises Electronic Health Records collected from a private hospital in Indonesia. The dataset encompasses samples from $3309$ patients; each sample consists of 8 scalar columns representing $8$ kinds of continuous-valued laboratory blood test results and 2 binary variables, \texttt{SEX} and \texttt{SOURCE} which respectively represent our demographic and diagnosis variables $s$ and $y$. 

Unlike most EHR datasets, this set was by default reasonably balanced, making it an optimal choice for testing our methods. To ensure the dataset was exactly balanced at the start of our experiment, we performed random undersampling such that each value of \texttt{SEX} was represented equally. Since the dataset was already nearly balanced, this step only discarded a handful of samples. We then generated a train/test split, where the train set serves as a ``best-case" control (referred to as the ``original" set) and a test set provides an equitable set for our experiments. Next, we deliberately created an imbalanced dataset by randomly drawing only $10\%$ of samples from the minority class \texttt{F}, and $100\%$ of samples from the majority class \texttt{M}. After creating the imbalanced datasets, the set the was used to train the CDDPM contained $1792$ samples. 

DDPM models have historically exhibited optimal performance with high dimensional datasets, such as those found in images, video, and sound. This dataset would normally be considered poor for the application of such models due to its low dimensionality, limited number of samples, and lack of translation or chirality invariances in our dataset when selecting it. However, these same traits make the set a good adversarial test set for MCRAGE. Ultimately, our method showcased effectiveness even on this maladapted dataset, implying potential success in a majority of real-world applications.

\subsection{Experimental Setup}

Our experiment consisted of two control and two treatment groups. Our control groups are the original and imbalanced datasets. We applied MCRAGE and SMOTE as our two treatment groups. 

The more complex and time-consuming treatment was the MCRAGE group. In order to tune the CDDPM model involved in MCRAGE, we first found a $\beta$-schedule which worked, then fixed the diffusion time complexity $T$, and finally performed a grid search of $25$ settings for learning rate and dropout rate. We trained each model instance using the value $T' = 35$ for $10000$ epochs, and saved the best checkpointed model. Every $100$ epochs, we crossvalidated the model by sampling the model according to MCRAGE and testing the performance of a classifier trained on that set. In order to select a best model checkpoint, we selected the diffusion model which generated the set that trained the classifier which achieved the highest F1 score on a $10\%$ validation split taken from the Imbalanced dataset. The best model was reloaded and was used to generate synthetic minority data, which was then concatenated to the original data according to the MCRAGE algorithm.

The SMOTE treatment group was simple, we applied SMOTE to the data, using labels identical to $s_0$ in the MCRAGE algorithm. The SMOTE was significantly simpler and less computationally expensive, but nonetheless offered a useful comparison for MCRAGE.

\begin{figure}
    \centering
    \includegraphics[width = .93\linewidth]{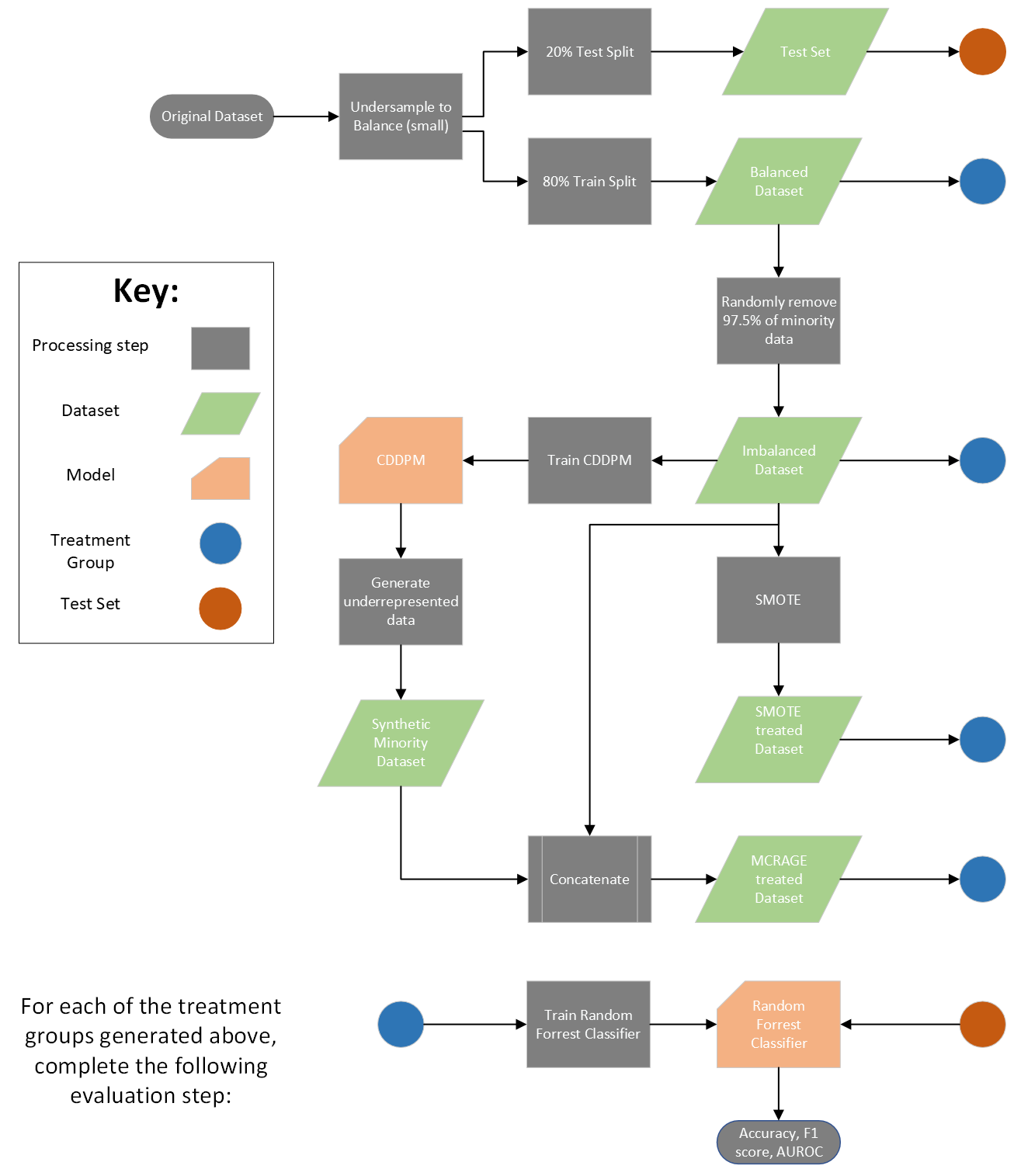}
    \caption{Flowchart detailing the experimental procedure}
    \label{fig:flowchart}
\end{figure}

After each treatment dataset was created, it was used to train a Random Forest Classifier, which was then tested on the test set. We report the resultant Accuracy, F1, and AUROC for the classifier trained on data in each of the treatment groups. We have provided a flowchart (Figure \ref{fig:flowchart}) for the readers convenience in understanding our setup.

\subsection{Sample Quality and Rebalancing Evaluation}

\begin{figure}
  \centering
  \begin{subfigure}{0.4\textwidth}
    \includegraphics[width=\linewidth]{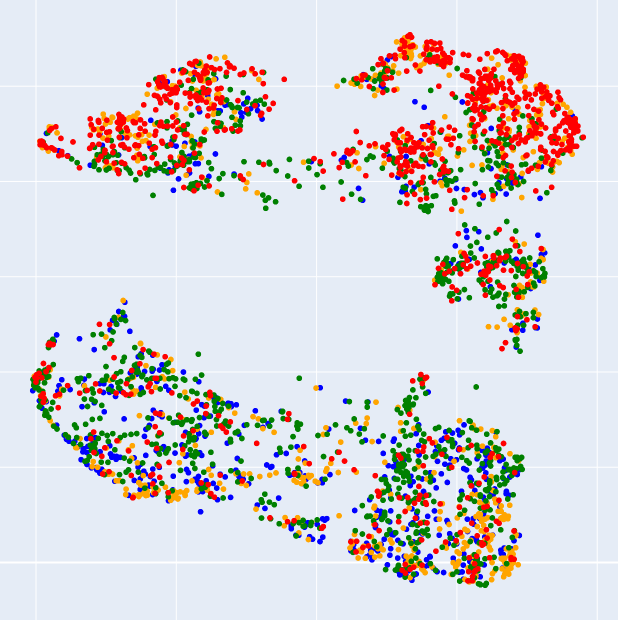}
    \caption{Balanced Dataset}
    \label{fig:image1}
  \end{subfigure}
  \hspace{1.5cm} 
  \begin{subfigure}{0.4\textwidth}
    \includegraphics[width=\linewidth]{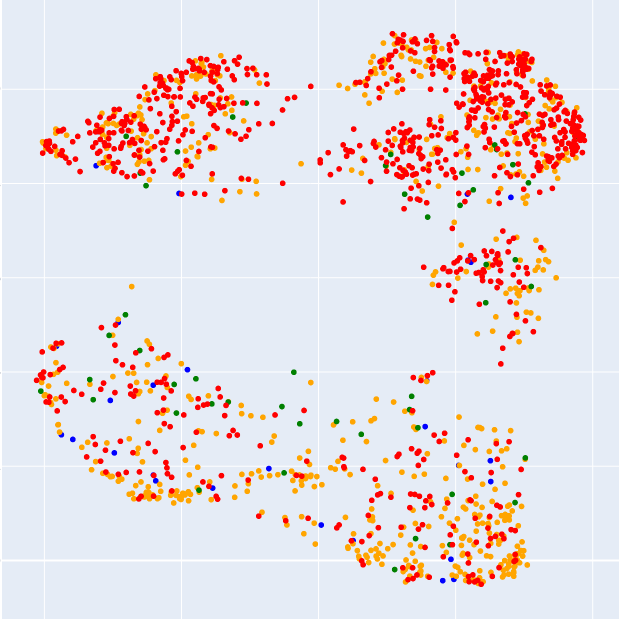}
    \caption{Imbalanced Dataset}
    \label{fig:image2}
  \end{subfigure}

  \vspace{0.3cm} 

  \begin{subfigure}{0.4\textwidth}
    \includegraphics[width=\linewidth]{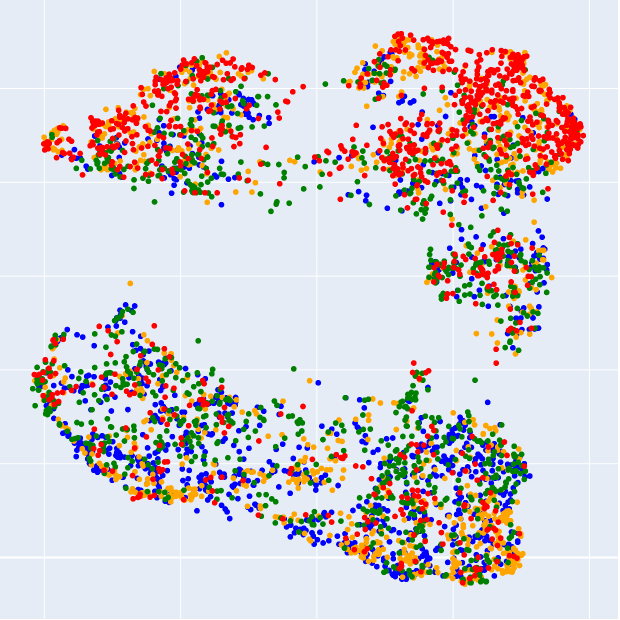}
    \caption{SMOTE-treated Dataset}
    \label{fig:image3}
  \end{subfigure}
  \hspace{1.5cm} 
  \begin{subfigure}{0.4\textwidth}
    \includegraphics[width=\linewidth]{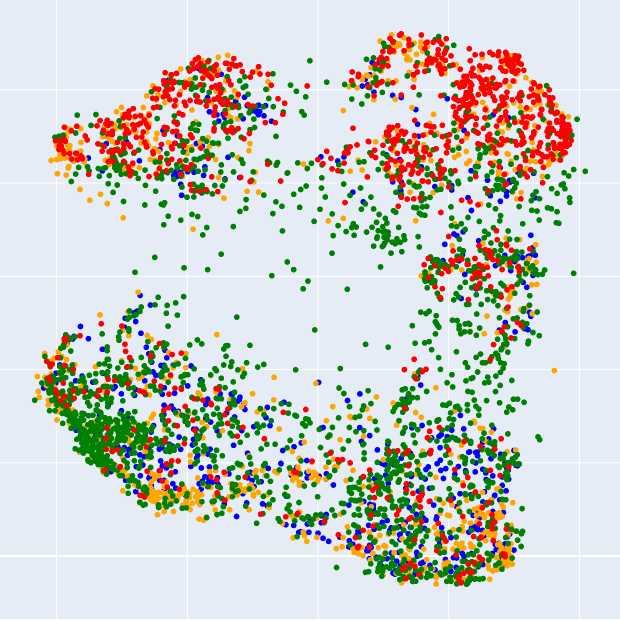}
    \caption{MCRAGE-treated Dataset}
    \label{fig:image4}
  \end{subfigure}
  \caption{Manifold Projections of Classifier Training Datasets}
  \label{fig:grid}
\end{figure}

In order to verify that the generated samples were meeting our expectations in terms of fidelity, we needed a method of easily and subjectively assessing sample quality. For this purpose, we used UMAP to generate manifold projections of our synthetic datasets and compared them to the original balanced and artificially imbalanced sets \cite{umap_docs}. Among the plots in Figure \ref{fig:grid}, it is evident that the MCRAGE treated set is qualitatively more similar to the balanced set than the alternative SMOTE-treated set. In our setting, where the primary concern is the performance of downstream classifiers, the SMOTE method fails to generalize the trend of the minority data. 

In particular, SMOTE is an inherently interpolation-based method, meaning that all samples generated by the technique are inside the convex hull of the original minority data. In practice, when the minority group is sparse, SMOTE results in isolated clusters of minority samples that do not have enough variance for a classifier trained on SMOTE-treated data to adequately generalize many decision boundaries. This is detrimental to our goal of improving classifier performance, as the resultant minority samples must have sufficient variance for the model to adequately learn a decision boundary that will perform well when diagnosing individuals in the minority class.

As an empirical investigation of the theoretical convergence of CDDPMs, we sampled 4000 points from each class using our tuned CDDPM model and plotted the distributions against the original data (Figure \ref{fig:hist}). The resulting histograms seem to indicate that the conditioned samples are in fact converging to the conditional distribution represented in the data.

\subsection{Classifier Fairness Evaluation}
\begin{figure}
  \centering
  \begin{subfigure}{1\textwidth}
    \includegraphics[width=\linewidth]{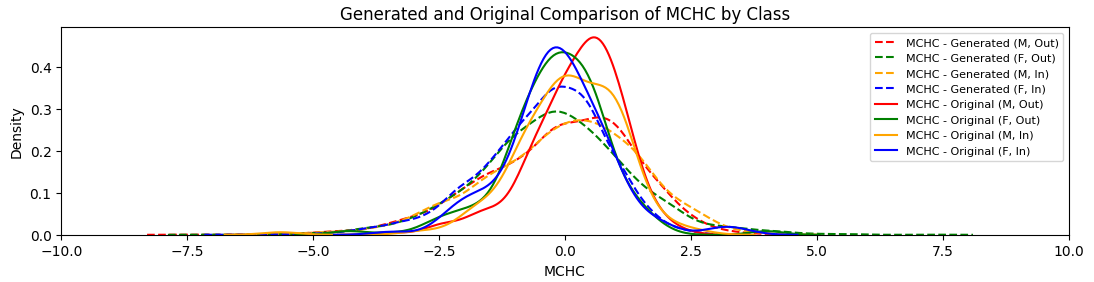}
    \label{fig:image5}
  \end{subfigure}
  \begin{subfigure}{1\textwidth}
    \includegraphics[width=\linewidth]{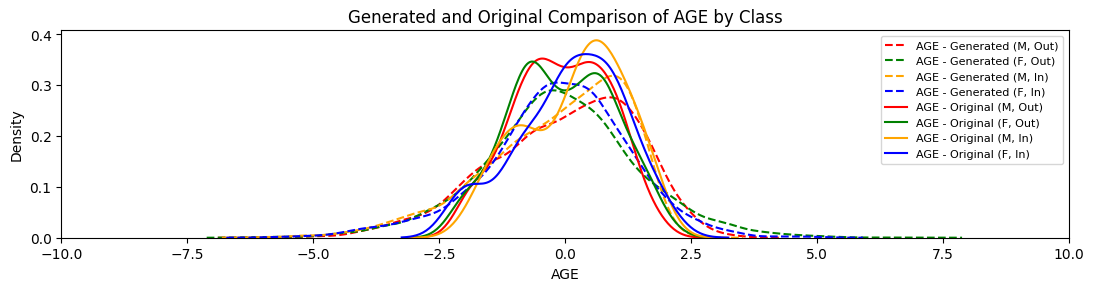}
    \label{fig:image6}
  \end{subfigure}
  \caption{Conditional Sampling Distributions as compared to Data Distributions for 2 selected features of 9.}
  \label{fig:hist}
\end{figure}

We will demonstrate the utility of our method with a binary classification task using a Random Forest classifier. For comparison to the current state-of-the-art, we also use SMOTE to rebalance the imbalanced dataset. Then, we evaluate the performance of the random forest classifier on each of the treated datasets and the balanced and imbalanced control sets described previously. Resulting metrics are shown in the Table 1. 
To evaluate the effectiveness of DDPM augmentation in improving downstream classifier fairness, we assess F1 score, which is the harmonic mean of precision and recall, because it considers both false positives and false negatives, making it more robust to class imbalances because it gives equal weight to both types of errors. In line with these assessments, we plotted kernel density estimation (KDE) plots, as shown in Figure \ref{fig:hist}, that compare the distributions of generated data against original data for selected features, to validate that our generated data distribution approximates the true distribution well. 

Our method shows a clear improvement over both the imbalanced and SMOTE-treated datasets. As seen in Table \ref{tab:results_table}, the MCRAGE treated classifier shows a $4.69\%$ increase in F1 score over the imbalanced classifier, and a $4.42\%$ increase over that of the SMOTE-treated classifier. As expected, the SMOTE-treated classifier shows a modest accuracy loss of $1.11\%$ over the imbalanced control, whereas the MCRAGE treated one gained $1.59\%$. This potentially confirms our earlier observation that SMOTE tends to overfit, leading to potential losses in test performance. Surprisingly, the MCRAGE group classifier recieves a $2.80\%$ increase in F1 score versus the balanced control group. This defies conventional intuition because, treating the MCRAGE process as one model, that model has a much worse training set than the balanced control set. However, the balanced control set is not \textit{intersectionally} balanced, so the classifier trained on this set may still have discrepancies in performance between intersectional groups, leading to lower F1-score. Overall, the MCRAGE treated classifier exceeded expectations in terms of F1 performance, demonstrating its novel utility as a dataset preprocessing step to promote fairness in downstream classifiers.

Moreover, as seen in Figure \ref{fig:hist}, for the features ``MCHC" and ``AGE", our generated data distributions closely match the original distributions. This serves as motivation for future work in proving theoretical convergence results for conditioned diffusion models. This experiment verifies that the MCRAGE process can reliably increase the fairness of downstream classifiers relative to no treatment of demographic imbalance or SMOTE.


\begin{table} [H]
    \centering
        \begin{tabular}{|c|c|c|c|c|}
            \hline
            & Imbalanced & SMOTE & MCRAGE & Balanced \\
            \hline 
            Accuracy (\%) & 71.348 & 70.555  & 72.480 & 73.160 \\
            \hline
            F1 Score & 0.64215 & 0.64384 & \textbf{0.67228} & 0.65396 \\
            \hline
            AUROC & 0.70 & 0.70 & 0.72 & 0.71 \\
            \hline
        \end{tabular}
    \caption{Results of random forest classifier trained on different datasets.}
    \label{tab:results_table}
\end{table}

\section{Discussion of Results}

The numerical experiment shows that MCRAGE treated data yields superior results in training fair downstream classifiers compared to the same process implemented with SMOTE treated or Imbalanced dataset. Our method yields significant improvement in accuracy, F1 score, and AUROC, where out of all the models only the balanced control classifier outperformed the MCRAGE treated one, and even then only in terms of raw accuracy. This demonstrates a novel application of the CDDPM architecture to promote fairness in healthcare or other consequential classification tasks.

In practice, most EHR datasets will perform like our imbalanced set due to intersectional imbalances, and the balanced set will be inaccessible. In situations where there is a significant class imbalance, it is beneficial to apply synthetic minority sampling techniques. There are cases where SMOTE may not be as effective, however: in datasets with sparse minority groups, SMOTE-generated samples may exhibit a cluster-like behavior, so the synthetic samples generated by SMOTE may be concentrated in certain regions of the feature space, leading to potential losses in classification performance.

Implementing the MCRAGE process involves choosing an appropriate CDDPM architecture, tuning several hyperparameters, and often many training runs before achieving usable results. In practice, obtaining a model which can generate quality samples requires substantial time, computational resources, and significant patience. By contrast, SMOTE is relatively simple and only has one parameter $k$, the number of neighbors to sample. We justify the difference in implementation cost by the generality of application, broad evidence of performance, and explainability of fairness by way of the theoretical convergence guarantees stated above. In certain applications such as automated healthcare, benefits such as generality, performance, and explainability are simply worth this additional cost.

\label{sec:conclusions}
\section{Future Work and Limitations}
The MCRAGE algorithm, presented in this work, represents a significant advancement in treating the pervasive issue of classifier bias stemming from demographic under-representation in training data. While this project has focused on applications to healthcare, similar methods could be applied to many other demographically-sensitive data; MCRAGE's rigorous and versatile framework make it applicable across many fields.

To enhance the practicality and efficiency of MCRAGE, we propose a future approach that could further optimize the method's performance. In practice, applying SMOTE to the data used to subsequently train a CDDPM seems be the best strategy. By training the generative model on a dataset containing additional interpolated minority samples, the model is given more information and thus seems to obtain even better convergence for those classes. Although this method generated an exceptional F1 score, this method may not generalize as well, since the CDDPM will converge to a distribution which has been corrupted by SMOTE. This approach offers a promising general-purpose fairness preprocessing step for demographic disparity in data, and future testing may determine if it is as reliable as stock MCRAGE.

The field of generative modeling is characterized by dynamic advancements, and continuous improvements in generative model architectures are expected to lead to more robust and efficient results. Investigating similar architectures such as Mixtures of Experts of CDDPM, CDDPM with different class guidance, conditional Poisson Flow Generative Models (PFGM) \cite{xu2022poisson} may deliver better samples and thus improve the performance of the process. These innovations can offer more efficient and equally effective solutions, further establishing MCRAGE as a pioneering approach in healthcare AI.

In conclusion, MCRAGE promises to mitigate data-induced classifier bias in healthcare AI using a standardized framework for fairness-motivated synthetic data methods. By guaranteeing a best estimate of an equitable sample at relatively low cost, MCRAGE ensures that equitable healthcare outcomes are available regardless of patient demographics or model architecture.

\section*{Acknowledgments}
This work is sponsored by NSF DMS 2051019. We would like to thank Dr. Yuanzhe Xi for his mentorship during the REU program. We would also like to acknowledge Huan He and Tianshi Xu for their collaboration and insight.

\bibliographystyle{siamplain}
\bibliography{references}

\end{document}